\documentclass[sigconf]{acmart}

\usepackage{booktabs} 
\usepackage[english]{babel}
\usepackage{color}
\usepackage{graphicx}
\usepackage[utf8]{inputenc}
\usepackage{multirow}
\usepackage{rotating}
\usepackage[export]{adjustbox}

\newcommand{\sdag}{$^{\dagger}$}
\newcommand{\ssdag}{$^{\dagger\dagger}$}
\newcommand{\spsdag}{$^{\dagger}$\phantom{$^{\dagger}$}}
\newcommand{\psdag}{\phantom{$^{\dagger}$}}
\newcommand{\pssdag}{\phantom{$^{\dagger\dagger}$}}

\begin{document}

\copyrightyear{2018} 
\acmYear{2018} 
\setcopyright{licensedothergov}
\acmConference[CIKM '18]{The 27th ACM International Conference on Information and Knowledge Management}{October 22--26, 2018}{Torino, Italy}
\acmBooktitle{The 27th ACM International Conference on Information and Knowledge Management (CIKM '18), October 22--26, 2018, Torino, Italy}
\acmPrice{15.00}
\acmDOI{10.1145/3269206.3269287}
\acmISBN{978-1-4503-6014-2/18/10}

\title{A Recurrent Neural Network for Sentiment Quantification}


\author{Andrea Esuli, Alejandro Moreo Fernández, Fabrizio Sebastiani}
\affiliation{Istituto di Scienza e Tecnologie dell'Informazione}  
\affiliation{Consiglio Nazionale delle Ricerche}
\affiliation{56124 Pisa, Italy}
\email{firstname.lastname@isti.cnr.it}


\begin{abstract}
  \emph{Quantification} is a supervised learning task that consists in
  predicting, given a set of classes $\mathcal{C}$ and a set $D$ of
  unlabelled items, the prevalence (or relative frequency) $p_{c}(D)$ of each class
  $c\in\mathcal{C}$ in $D$. Quantification can in principle be solved
  by classifying all the unlabelled items and counting how many of
  them have been attributed to each class. However, this ``classify
  and count'' approach has been shown to yield suboptimal
  quantification accuracy; this has established quantification as a
  task of its own, and given rise to a number of methods specifically
  devised for it.  
  We propose a recurrent neural network architecture for quantification (that we call
  \emph{QuaNet}) that observes the classification predictions to learn 
  higher-order ``quantification embeddings'', which are then refined by incorporating
  quantification predictions of simple classify-and-count-like
  methods. 
  We test \emph{QuaNet} on sentiment quantification on
  text, showing that it substantially outperforms several
  state-of-the-art baselines.
\end{abstract}


\maketitle


\section{Introduction}
\label{sec:intro}

\noindent \emph{Quantification} (also known as ``class prior
estimation'') is a supervised learning task that consists of
predicting, given a set of classes $\mathcal{C}$ and a set $D$ (a
\emph{sample}) of unlabelled items drawn from some domain
$\mathcal{D}$, the prevalence (i.e., relative frequency) $p_{c}(D)$ of
each class $c\in\mathcal{C}$ in $D$. Put it another way, given an
unknown distribution $p_{\mathcal{C}}(D)$ of the members of $D$ across
$\mathcal{C}$ (the \emph{true distribution}), quantification consists
in generating a \emph{predicted distribution}
$\hat{p}_{\mathcal{C}}(D)$ that approximates $p_{\mathcal{C}}(D)$ as
accurately as possible \cite{Gonzalez:2017it}. Quantification is
important for many application fields characterised by an interest in
aggregate (rather than individual) data, such as the social sciences,
market research, political science, and epidemiology.

In principle, quantification can be trivially solved via
classification, i.e., by training a classifier $h$ using training data
labelled according to $\mathcal{C}$, classifying the unlabelled data
in $D$ via $h$, and counting, for each $c\in\mathcal{C}$, how many
items in $D$ have been attributed to $c$ (the ``classify and count''
method). However, research has conclusively shown that this approach
leads to suboptimal quantification accuracy \cite{Barranquero:2015fr,Bella:2010kx,Esuli:2015gh,Forman:2008kx}. 
Simply
put, the reason is that classifiers are typically trained to minimize
classification error, which is by and large proportional to $(FP+FN)$,
while a good quantifier should be trained to minimise quantification
error, which is by and large proportional to $|FP-FN|$ (where $TP$, $FP$, $FN$, $TN$ denote the usual counts
from a binary contingency table). 

In this paper we tackle quantification in a binary setting, and
propose a recurrent neural network architecture (that we call
\emph{QuaNet}) that observes the classification predictions to learn higher-order ``quantification embeddings'', which are then refined by incorporating
quantification predictions of simple classify-and-count-like
methods.
We apply this method to the problem of performing quantification by
sentiment (i.e.,
$\mathcal{C}$=\{\textsc{Positive},\textsc{Negative}\}) on text. The aspect of sentiment is of special interest for quantification
\cite{Esuli:2010fk,Gao:2016uq}, since quantification (as
applied to e.g., posts from social networks, or product reviews) is a key tool to monitor
opinion trends on objects of interest (e.g., products, policies, or
political candidates).

To the best of our knowledge, QuaNet is the first deep learning
approach to sentiment quantification proposed in the literature.


\section{Related Work}
\label{sec:related}

\begin{figure*}
  \includegraphics[width=0.9\textwidth]{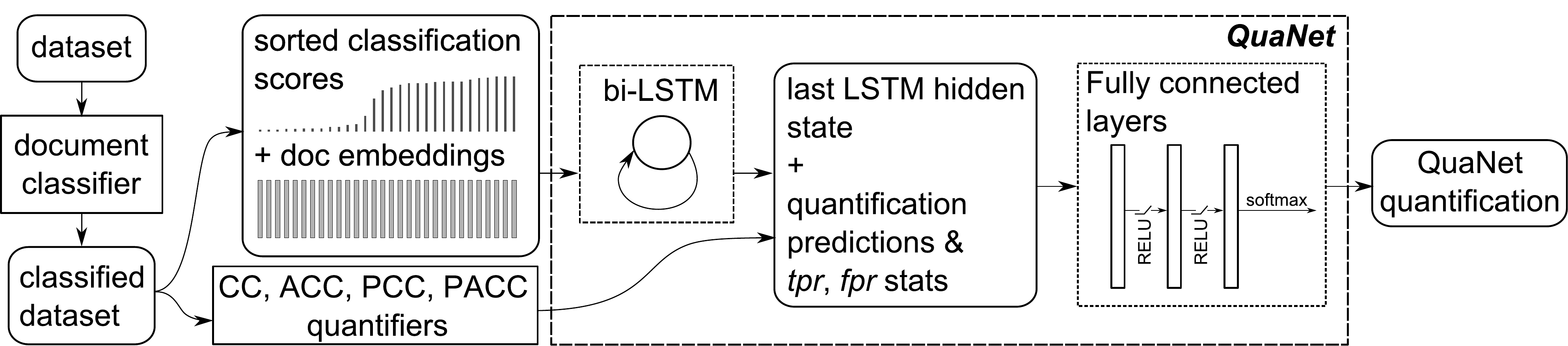}
  \caption{\label{fig:quanet} Architecture of the QuaNet
  quantification system.}
\end{figure*}

\noindent The simplest quantification method (and the one that acts as
a lower-bound baseline for all quantification methods) is the
above-mentioned \emph{classify and count} (CC), which comes down to
computing
\begin{equation}\label{eq:cc}
  \begin{aligned}
    \hat{p}_{c}^{CC}(D) & = \frac{|\{\mathbf{x}\in
    D|h(\mathbf{x})=c\}|}{|D|} = \frac{TP_{b}+FP_{b}}{|D|}
  \end{aligned}
\end{equation}

\noindent where the $b$ subscript indicates that the 
values are counts of \underline{b}inary predictions (the reason for
specifying this will become clear later).  It is easy to verify that
the relation between $\hat{p}_{c}^{CC}(D)$ and the true prevalence
$p_{c}(D)$ is given, in the binary case,\footnote{Equation
\ref{eq:exactacc} is easily extended to the multiclass case (see e.g.,
\cite{Gao:2016uq}). For brevity, in the rest of the paper we will
exclusively deal with the binary case.}  by
\begin{equation}
  \label{eq:exactacc} 
  p_{c}(D) = \frac{\hat{p}^{CC}_c(D) - fpr_{b}}{tpr_{b} - fpr_{b}}
\end{equation}

\noindent where $tpr_{b}$ and $fpr_{b}$ denote the \emph{true positive
rate} $tpr_{b}=\frac{TP_{b}}{TP_{b}+FN_{b}}$ and \emph{false positive
rate} $fpr_{b}=\frac{FP_{b}}{FP_{b}+TN_{b}}$ of $h$ on $D$. This
relationship justifies the \emph{adjusted classify and count} (ACC)
quantification method \cite{Forman:2008kx}, which consists in
computing
\begin{equation}
  \label{eq:acc} 
  \hat{p}_{c}^{ACC}(D) = \frac{\hat{p}^{CC}_c(D) - \hat{fpr}_{b}}{\hat{tpr}_{b}-\hat{fpr}_{b}}
\end{equation}

\noindent where $\hat{tpr}_{b}$ and $\hat{fpr}_{b}$ are estimates of
$tpr_{b}$ and $fpr_{b}$ computed either on a held-out set or via $k$-fold cross-validation. 

CC and ACC use the binary predictions generated by classifier
$h$. Since most classifiers can be configured to output ``soft
predictions'' in the form of posterior probabilities
$\Pr(c|\mathbf{x})$ (from which binary predictions are obtained via
thresholding),\footnote{If a classifier natively outputs
classification scores that are not probabilities, the former can be
converted into the latter via ``probability calibration''; see
e.g., \cite{Platt:2000fk}.} and since  posterior probabilities contain richer information than binary predictions, one can generate probabilistic versions of the CC and
ACC methods \cite{Bella:2010kx} based on ``soft contingency tables'',
i.e., contingency tables where counts $TP_{b}$, $FP_{b}$, $FN_{b}$,
$TN_{b}$ are replaced by \emph{expected counts} $TP_{s}$, $FP_{s}$,
$FN_{s}$, $TN_{s}$ (subscript $s$ indicating that the corresponding
values derive from \underline{s}oft predictions) obtained from the

\begin{center}
  \begin{tabular}{|c|c|l|l|}
    \hline
    \multicolumn{2}{|c|}{\mbox{}} & \multicolumn{2}{|c|}{True} \\
    \cline{3-4}
    \multicolumn{2}{|c|}{\mbox{}} & 
                                    \multicolumn{1}{|c}{\textsc{Pos}} & \multicolumn{1}{|c|}{\textsc{Neg}} \\
    \hline
    \rule[-4ex]{0mm}{8ex}
    \multirow{2}{*}[-0.5em]{\begin{sideways}Predicted\end{sideways}} & \multirow{1}{*}{\begin{sideways}\textsc{Pos}\end{sideways}} & $TP_{s} = \displaystyle\sum_{\mathbf{x}\in c,D}\Pr(c|\mathbf{x})$ &                                                                                                                                                                                                 $FP_{s} = \displaystyle\sum_{\mathbf{x}\in \overline{c},D}\Pr(c|\mathbf{x})$  \\
    \cline{2-4}
    \rule[-4ex]{0mm}{8ex} & \multirow{1}{*}{\begin{sideways}\textsc{Neg}\end{sideways}} & $FN_{s} = \displaystyle\sum_{\mathbf{x}\in c,D}(1-\Pr(c|\mathbf{x}))$ &                                                                                                                                                                          $TN_{s} = \displaystyle\sum_{\mathbf{x}\in \overline{c},D}(1-\Pr(c|\mathbf{x}))$  \\
    \hline
  \end{tabular}
\end{center}
\bigskip
\noindent It is thus possible to define ``soft'' variants of $tpr_{b}$
and $fpr_{b}$ as $tpr_{s}=\frac{TP_{s}}{TP_{s}+FN_{s}}$ and
$fpr_{s}=\frac{FP_{s}}{FP_{s}+TN_{s}}$. One can then define
probabilistic versions of the quantification methods defined by
Equations \ref{eq:cc} and \ref{eq:acc}, namely, \emph{probabilistic
classify and count} (PCC), defined by
\begin{equation}\label{eq:pcc}
  \begin{aligned}
    \hat{p}_{c}^{PCC}(D) & = \frac{\sum_{\mathbf{x}\in
    D}\Pr(c|\mathbf{x})}{|D|} = \frac{TP_{s}+FP_{s}}{|D|}
  \end{aligned}
\end{equation}
\noindent and \emph{probabilistic adjusted classify and count} (PACC),
defined by
\begin{equation}
  \label{eq:pacc} 
  \hat{p}_{c}^{PACC}(D) = \frac{\hat{p}^{PCC}_c(D) - \hat{fpr}_{s}}{\hat{tpr}_{s}-\hat{fpr}_{s}}
\end{equation}
 
\noindent The four quantification methods defined by Equations \ref{eq:cc},
\ref{eq:acc}, \ref{eq:pcc}, \ref{eq:pacc}, use the output of
general-purpose classifiers. A radically different approach is taken
in \cite{Esuli:2015gh}, which suggests to apply Equation \ref{eq:cc} to the
output of a classifier specifically designed for optimizing
quantification accuracy. Since any measure of quantification accuracy
is inherently nonlinear and multivariate, \cite{Esuli:2015gh} suggests
using the $SVM_{perf}$ structured output learner \cite{Joachims05}, which can indeed optimize any
nonlinear and multivariate loss function that can be expressed in
terms of a contingency table. Three quantification methods have
emerged in the literature as a result of this proposal, namely, (i)
SVM$^{KLD}$ \cite{Esuli:2015gh}, which minimizes the
\emph{Kullback-Leibler divergence} ($KLD$ -- see Equation
\ref{eq:kld}), (ii) SVM$^{NKLD}$ \cite{Esuli:2014uq}, which minimizes a
normalized version of $KLD$,
and (iii) SVM$^{Q}$ \cite{Barranquero:2015fr}, which minimizes the
harmonic mean of a classification-oriented loss (recall) and a
quantification-oriented loss ($RAE$ -- see Equation \ref{eq:rae}). 
These methods still represent the state of the art in terms of 
quantification effectiveness.

Yet another approach to quantification is exemplified by the EMQ
method of \cite{Saerens:2002uq}, which consists in generating a
probabilistic classifier while exploiting the EM algorithm to
iteratively shift the estimation of $p_{\mathcal{C}}(D)$ from the one
of the training set to one that maximizes the likelihood on the test
data.

In Section \ref{sec:experiments} we will use all the quantification
methods presented in this section as baselines against which to
 compare our approach.


\section{Deep Learning for Quantification}
\label{sec:method}

\noindent
We propose a new deep learning method for quantification, that we
dub QuaNet.  The QuaNet architecture (see Figure
\ref{fig:quanet}) is composed of a main
recurrent component (a bidirectional LSTM) that, given a class $c$ and
a classifier $h$ which returns posterior probabilities
$\Pr(c|\mathbf{x})$, receives as input a list $L$ of pairs
$\langle\Pr(c|\mathbf{x}),\vec{\mathbf{x}}\rangle$, one for each
$\mathbf{x}\in D$. Here, $\vec{\mathbf{x}}$ is the document
embedding for $\mathbf{x}$, and $L$ is sorted by the value of
$\Pr(c|\mathbf{x})$.  The intuition is that the recurrent network
``learns to count'' the \textsc{Positive} and \textsc{Negative} examples by observing
them.  The rationale of \emph{sorting} the list of posterior probabilities is that by
observing the ordered sequence of $\Pr(c|\mathbf{x})$
values the network can learn to recognize the switch point between
\textsc{Negative} and \textsc{Positive} documents.
The document embedding $\vec{\mathbf{x}}$ associated to each
score is a vectorial representation of the content of the document.  The embedding can be exploited by the network for
learning to weight the contribution of each document to the final
estimate.

The vector returned by the LSTM can be considered a ``quantification embedding'', i.e., a dense, multi-dimensional representation of the information  relevant to quantification observed from the input data. This vector is then concatenated to a short vector
that contains (i) the $\hat{p}_{c}^{CC}(D)$, $\hat{p}_{c}^{ACC}(D)$, $\hat{p}_{c}^{PCC}(D)$ and $\hat{p}_{c}^{PACC}(D)$ predictions, and (ii) the $tpr_{b}$,
$fpr_{b}$, $tpr_{s}$,
$fpr_{s}$ statistics, i.e., a set of
aggregate values that are very cheap to determine from the classifier $h$.\footnote{In additional experiments, not reported in this paper for reasons of space, we have performed a number of ablation experiments which indeed confirmed that all these inputs to QuaNet do contribute to improving the quantification accuracy of QuaNet.}  
The resulting vector passes through the second
component of QuaNet, that consists of a number of fully connected
layers, with ReLU activation, which ``correct'' the quantification embedding from the LSTM by exploiting the explicit quantification-related statistics received from the classifier. The final quantification embedding from this stage is then converted into the $\hat{p}^{QuaNet}_c(D)$ prediction by a final layer of size 2 with softmax activation.

In principle, QuaNet could exploit (along those coming from CC, ACC, PCC, PACC) quantification predictions from any
method, and the richer the set of predictions the more information the
network would have for returning an accurate output.  However, it is reasonable to use
only quantification predictions that can be computed with minimal
computational cost;
for this reason we stick to
just the above mentioned predictions that are
computed directly from the output of  classifier $h$, while the
$tpr_{b}$,
$fpr_{b}$, $tpr_{s}$,
$fpr_{s}$  
values needed for computing  $\hat{p}_{c}^{ACC}(D)$ and $\hat{p}_{c}^{PACC}(D)$  are easily determined by applying $h$ to a validation set.

In our QuaNet implementation\footnote{The implementations of QuaNet and of all the baseline methods
are
available at \url{https://github.com/HLT-ISTI/QuaNet}
} the LSTM cells have
64 hidden dimensions; the output of the LSTM (a vector of size 128, since the LSTM is bidirectional) is concatenated with the eight
quantification statistics from the classifier to form a vector with 136 dimensions that goes through two
dense layers of 1,024 and 512 dimensions, each with a ReLU activation
and a 0.5 dropout layer, ending with a layer of size two with softmax activation.


\section{Experiments}
\label{sec:experiments}


\noindent We run experiments on
three sentiment classification datasets: (i) \textsc{IMDB}, the
popular \emph{Large Movie Review Dataset} \cite{Maas2011}; (ii) \textsc{Kindle}, a set of reviews of Kindle
e-book readers, and (iii)
\textsc{HP}, a set of reviews of the books from the Harry Potter
series. The \textsc{IMDB} dataset consists of 25,000 training and
25,000 test documents. We crawled the \textsc{Kindle} and \textsc{HP} datasets  
ourselves from the Amazon
website; we consider all 5-stars and 4-stars reviews as \textsc{Positive} and all 1-star and 2-stars reviews as \textsc{Negative}, discarding all 3-stars reviews.\footnote{These datasets
are available at \url{http://hlt.isti.cnr.it/quantification/}} The
\textsc{Kindle} dataset consists of the 25,413 reviews from Aug 2008 to Jul 2011;
we take the 3,821 reviews from the first three months as the training
set, leaving 21,592 reviews for testing. The HP corpus consists of the 27,934 reviews from 1998 to 2011; we
use the 9,533 reviews from the first three years for
training and the remaining 18,401 documents for testing. 
From each set of training data we randomly select 60\% for training purposes, leaving the remaining 40\% for parameter optimization.
The \textsc{IMDB}, \textsc{Kindle}, and \textsc{HP} datasets are examples of balanced, imbalanced, and severely imbalanced datasets, since their training set prevalence of the \textsc{Positive} class is 50.0\%, 91.7\% 98.2\%, respectively.

\begin{figure*}
  \includegraphics[width=0.98\textwidth]{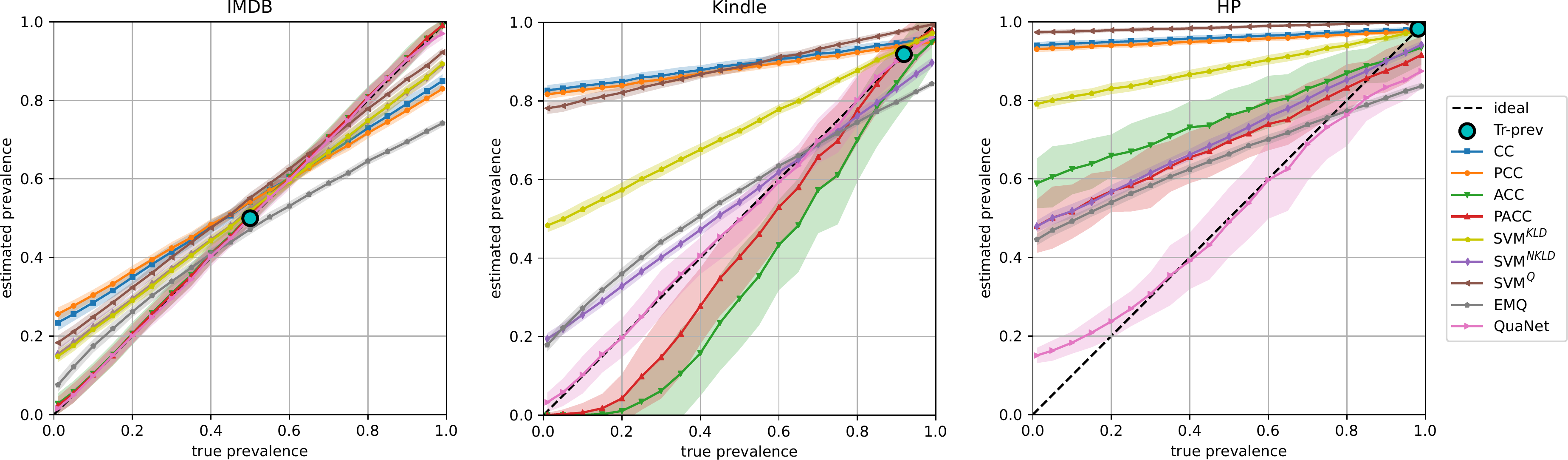}
  \caption{\label{fig:results} Prevalences of the \textsc{Positive} class predicted by the various
  methods, against the corresponding true prevalences.
  }
\end{figure*}

\begin{table*}[ht!]
  \resizebox{\textwidth}{!} {
  \begin{tabular}{|l||rr|rr|rr||rr|rr|rr||rr|rr|rr|}
    \hline
     & \multicolumn{6}{c||}{\textsc{IMDB}} & \multicolumn{6}{|c||}{\textsc{Kindle}} & \multicolumn{6}{|c|}{\textsc{HP}} \\\hline
    & \multicolumn{2}{c}{AE} & \multicolumn{2}{|c}{RAE} & \multicolumn{2}{|c||}{KLD} & \multicolumn{2}{|c}{AE} & \multicolumn{2}{|c}{RAE} & \multicolumn{2}{|c||}{KLD} & \multicolumn{2}{|c}{AE} & \multicolumn{2}{|c}{RAE} & \multicolumn{2}{|c|}{KLD} \\ \hline\hline
    CC       &        0.096\pssdag & (+421\%) &    1.193\psdag & (+1008\%) &  0.044 & (+1419\%) &     0.417 & (+585\%) &        5.805\pssdag & (+1083\%) &      0.717\psdag & (+3288\%) &  0.476 & (+379\%) &      6.487\psdag & (+526\%) &  1.346 & (+2789\%) \\
    ACC      &         0.021\pssdag & (+15\%) &      0.144\psdag & (+34\%) &   0.011 & (+273\%) &     0.160 & (+161\%) &      0.884\spsdag & (+80\%) &  0.839\sdag & (+3866\%) &  0.245 & (+146\%) &      2.919\psdag & (+181\%) &   0.400 & (+759\%) \\
    PCC      &        0.107\pssdag & (+483\%) &    1.336\psdag & (+1141\%) &  0.053 & (+1704\%) &     0.407 & (+567\%) &        5.704\pssdag & (+1062\%) &      0.661\psdag & (+3022\%) &  0.469 & (+371\%) &      6.411\psdag & (+519\%) &  1.202 & (+2480\%) \\
    PACC     &  0.019\ssdag & (+1\%) &  0.126\sdag & (+17\%) &   0.010 & (+233\%) &     0.138 & (+127\%) &  0.688\ssdag & (+40\%) &  0.736\sdag & (+3378\%) &  0.205 & (+106\%) &  2.391\sdag & (+130\%) &   0.267 & (+473\%) \\
    SVM$^{\mathit{KLD}}$  &        0.065\pssdag & (+253\%) &     0.890\psdag & (+727\%) &   0.025 & (+741\%) &     0.228 & (+273\%) &         3.186\pssdag & (+549\%) &       0.171\psdag & (+708\%) &  0.385 & (+287\%) &      5.298\psdag & (+411\%) &  0.543 & (+1065\%) \\
    SVM$^{\mathit{NKLD}}$ &        0.068\pssdag & (+269\%) &     0.938\psdag & (+771\%) &   0.026 & (+795\%) &  0.080 & (+32\%)\dag &         1.192\pssdag & (+142\%) &    0.033\sdag & (+58\%) &  0.217 & (+118\%) &      3.146\psdag & (+203\%) &   0.160 & (+244\%) \\
    SVM$^{\mathit{Q}}$    &        0.075\pssdag & (+305\%) &     1.127\psdag & (+947\%) &   0.029 & (+907\%) &     0.387 & (+535\%) &         5.246\pssdag & (+969\%) &      0.548\psdag & (+2489\%) &  0.486 & (+389\%) &      6.560\psdag & (+533\%) &  1.643 & (+3427\%) \\
    EMQ       &        0.094\pssdag & (+411\%) &     0.529\psdag & (+391\%) &  0.045 & (+1428\%) &      0.104 & (+70\%) &         1.196\pssdag & (+143\%) &       0.046\psdag & (+115\%) &  0.200 & (+101\%) &      2.912\psdag & (+181\%) &   0.140 & (+201\%) \\
    QuaNet  &          \textbf{0.018}\pssdag & &       \textbf{0.108}\psdag & &     \textbf{0.003} & &       \textbf{0.061} & &           \textbf{0.491}\pssdag & &         \textbf{0.021}\psdag & &    \textbf{0.099} & &        \textbf{1.036}\psdag & &     \textbf{0.047} & \\

    \hline
  \end{tabular}
  }
  \caption{Average errors of the quantification methods.
  Symbols $\dag$ and $\dag\dag$ denote the methods (if any) whose score is \emph{not} statistically significantly different from the best one  (in \textbf{boldface}) at $\alpha = 0.05$ ($\dag$) or at $\alpha = 0.005$ ($\dag\dag$) according to a two-tailed paired t-test measured on 10 runs of the experimental setup. Percentages indicate increase in error with respect to the best method.}
  \label{tab:results}
\end{table*}


In order to evaluate the effectiveness of a quantifier over
a wide spectrum of test prevalences we follow the experimental
protocol of \citet{Forman:2008kx}, which consists of randomly undersampling one of the two classes in the test set in order to obtain a sample with specified class prevalences. We extract samples with a prevalence of the \textsc{Positive} class in the set
$\{0.01, 0.05, 0.10, \ldots, 0.90, 0.95, 0.99\}$.\footnote{We do not
use 0 or 1 prevalences (and use instead 0.01 and 0.99). This is due to
the fact that, when $p_{c}(D)$ is 0 or 1, $RAE$ and $KLD$ (i) are undefined, and thus require smoothing to be defined, and even when smoothed, they are numerically unstable.}  For each of these 21 values we generate 100 random samples of 500 test documents each; we
thus report the averaged quantification error across
$21\times 100 = 2100$ test samples.  As the measures of quantification
error we use \emph{Absolute Error} ($AE$), \emph{Relative Absolute
Error} ($RAE$), and \emph{Kullback-Leibler Divergence}
($KLD$) \cite{Gonzalez:2017it}, defined as 
\begin{equation}
  \mathit{AE}(p,\hat{p})=\frac{1}{|\mathcal{C}|}\sum_{c\in \mathcal{C}}|\hat{p}(c)-p(c)| 
                            \label{eq:ae}
\end{equation}

\begin{equation}
  \mathit{RAE}(p,\hat{p})=\frac{1}{|\mathcal{C}|}\sum_{c\in 
                             \mathcal{C}}\displaystyle\frac{|\hat{p}(c)-p(c)|}{p(c)} 
                             \label{eq:rae}
\end{equation}
\begin{equation}
  \mathit{KLD}(p,\hat{p})=\sum_{c\in \mathcal{C}} 
                             p(c)\log_{e}\frac{p(c)}{\hat{p}(c)} 
                             \label{eq:kld} 
\end{equation}

\noindent We apply the above formulas individually on each test sample
and then take the mean across the resulting 2,100 values.
We compare our approach against all the methods discussed in Section \ref{sec:related}.

We adopt an LSTM network also as the document classifier\footnote{Note
that QuaNet is independent of the classifier, and thus the combination
of any classifier that outputs confidence scores with any method
producing document embeddings could in principle represent valid
inputs for QuaNet. Also, the classifier and QuaNet are trained and optimized separately.}
 since this
architecture naturally provides, for each input document, both a
classification score and an embedded representation as the last and
next-to-last layers, respectively.  In this LSTM each unique word
is first
transformed into a 100-dimensional embedding, which is then processed
by the LSTM cells with 128 hidden dimensions. The output of the LSTM
is followed by two fully connected layers (of 1,024 and 100 dimensions, resp.), each
followed by a ReLU activation and a 0.5 dropout layer. Another last
fully connected projection takes the 100-dimensional document embedding to
produce two outputs, via a softmax activation, which eventually
represent the confidence scores for the \textsc{Positive} and \textsc{Negative} classes.  

We use this LSTM-based classifier both 
for QuaNet and also for the CC, ACC, PCC, and PACC baselines, for a direct and fair comparison. For SVM$^{\mathit{KLD}}$, SVM$^{\mathit{NKLD}}$, and SVM$^{Q}$, we use Joachims' $SVM_{perf}$
implementation \cite{Joachims05}, instantiated with the
corresponding loss functions. For each of these methods we choose the $C$ parameter in the range
$C = 10^{i}$, with $i\in [-3,-2,\ldots 2, 3]$, that deliver the lowest
quantification error on the validation set.  As the base for EMQ we use the \texttt{scikit-learn}
implementation of Multinomial Na\"{i}ve
Bayes 
(MNB); we
choose the best value for its $\alpha$ parameter via grid search on
$\alpha=10^{i}$ with $i\in [-4,-2,\ldots 2, 4]$. 
As the loss for both neural networks (i.e., classifier and quantifier) we use the Mean Square Error ($MSE$) of the predictions, minimized via stochastic gradient descent; we apply the Adam updating rule, with learning rate $lr=10^{-4}$ and weight decay $wd=10^{-4}$. We set the batch size to 100 and the maximum number of iterations to 20,000, but we apply an early stopping criterion after 20 validations that do not show any improvement (we validate every 100 iterations).
Classification batches are composed of documents randomly sampled from the dataset.
Quantification batches contain quantification examples that uniformly cover the entire prevalence range.


\subsection{Results}
\label{sec:results}

\noindent Figure \ref{fig:results} displays the performance of the
different quantifiers.
Each point on a line denotes the average estimated prevalence across the 100 different random trials, while coloured
bands indicate the standard deviation. The pale blue circle indicates the training set prevalence of the \textsc{Positive} class; the (ideal) perfect quantifier is represented by the diagonal.

CC and PCC are adversely influenced by 
training class prevalences, and this is
especially true for the imbalanced and severely imbalanced \textsc{Kindle} and \textsc{HP}.  ACC and PACC  show
superior performance to their un-adjusted counterparts CC and PCC, as 
especially clear on IMDB;
yet they still leave room for improvement in the more difficult cases
of \textsc{Kindle} (where both tend to underestimate) and \textsc{HP} (where both tend to overestimate). 
The corrections tend to produce higher variances in the
estimations (as shown by the broader coloured bands). 
SVM-based methods are more stable,
producing very low variations across different samples; SVM$^{\mathit{NKLD}}$
shows better performance than SVM$^{\mathit{KLD}}$ and SVM$^{Q}$, especially on the \textsc{Kindle} and HP datasets. 
The plots show that QuaNet is consistently the best quantifier across all datasets.

Table \ref{tab:results} reports the quantification results for the
three evaluation metrics, averaged across 10 runs (the same 10
training/validation random splits are used consistently
across all methods).  
QuaNet produces much lower quantification error than any baseline, for all three error
measures and all three datasets.  PACC shows
comparable performance (in a statistically significant sense) to
QuaNet only in a few cases, e.g., on \textsc{IMDB} for $AE$ and in \textsc{Kindle} for
$RAE$.


\begin{acks}
  The first author thanks NVidia corp.\ for granting a Titan X GPU.
\end{acks}


\bibliographystyle{ACM-Reference-Format}

\begin{thebibliography}{12}


\ifx \showCODEN    \undefined \def \showCODEN     #1{\unskip}     \fi
\ifx \showDOI      \undefined \def \showDOI       #1{#1}\fi
\ifx \showISBNx    \undefined \def \showISBNx     #1{\unskip}     \fi
\ifx \showISBNxiii \undefined \def \showISBNxiii  #1{\unskip}     \fi
\ifx \showISSN     \undefined \def \showISSN      #1{\unskip}     \fi
\ifx \showLCCN     \undefined \def \showLCCN      #1{\unskip}     \fi
\ifx \shownote     \undefined \def \shownote      #1{#1}          \fi
\ifx \showarticletitle \undefined \def \showarticletitle #1{#1}   \fi
\ifx \showURL      \undefined \def \showURL       {\relax}        \fi
\providecommand\bibfield[2]{#2}
\providecommand\bibinfo[2]{#2}
\providecommand\natexlab[1]{#1}
\providecommand\showeprint[2][]{arXiv:#2}

\bibitem[\protect\citeauthoryear{Barranquero, D{\'\i}ez, and del
  Coz}{Barranquero et~al\mbox{.}}{2015}]%
        {Barranquero:2015fr}
\bibfield{author}{\bibinfo{person}{Jos{\'e} Barranquero},
  \bibinfo{person}{Jorge D{\'\i}ez}, {and} \bibinfo{person}{Juan~Jos{\'e} del
  Coz}.} \bibinfo{year}{2015}\natexlab{}.
\newblock \showarticletitle{Quantification-oriented learning based on reliable
  classifiers}.
\newblock \bibinfo{journal}{\emph{Pattern Recognition}} \bibinfo{volume}{48},
  \bibinfo{number}{2} (\bibinfo{year}{2015}), \bibinfo{pages}{591--604}.
\newblock
\urldef\tempurl%
\url{https://doi.org/10.1016/j.patcog.2014.07.032}
\showDOI{\tempurl}


\bibitem[\protect\citeauthoryear{Bella, Ferri, Hern{\'a}ndez-Orallo, and
  Ram{\'\i}rez-Quintana}{Bella et~al\mbox{.}}{2010}]%
        {Bella:2010kx}
\bibfield{author}{\bibinfo{person}{Antonio Bella}, \bibinfo{person}{C{\`e}sar
  Ferri}, \bibinfo{person}{Jos{\'e} Hern{\'a}ndez-Orallo}, {and}
  \bibinfo{person}{Mar{\'\i}a~Jos{\'e} Ram{\'\i}rez-Quintana}.}
  \bibinfo{year}{2010}\natexlab{}.
\newblock \showarticletitle{Quantification via Probability Estimators}. In
  \bibinfo{booktitle}{\emph{Proceedings of the 11th IEEE International
  Conference on Data Mining (ICDM 2010)}}. \bibinfo{address}{Sydney, {AU}},
  \bibinfo{pages}{737--742}.
\newblock
\urldef\tempurl%
\url{https://doi.org/10.1109/icdm.2010.75}
\showDOI{\tempurl}


\bibitem[\protect\citeauthoryear{Esuli and Sebastiani}{Esuli and
  Sebastiani}{2010}]%
        {Esuli:2010fk}
\bibfield{author}{\bibinfo{person}{Andrea Esuli} {and}
  \bibinfo{person}{Fabrizio Sebastiani}.} \bibinfo{year}{2010}\natexlab{}.
\newblock \showarticletitle{Sentiment quantification}.
\newblock \bibinfo{journal}{\emph{IEEE Intelligent Systems}}
  \bibinfo{volume}{25}, \bibinfo{number}{4} (\bibinfo{year}{2010}),
  \bibinfo{pages}{72--75}.
\newblock


\bibitem[\protect\citeauthoryear{Esuli and Sebastiani}{Esuli and
  Sebastiani}{2014}]%
        {Esuli:2014uq}
\bibfield{author}{\bibinfo{person}{Andrea Esuli} {and}
  \bibinfo{person}{Fabrizio Sebastiani}.} \bibinfo{year}{2014}\natexlab{}.
\newblock \showarticletitle{Explicit Loss Minimization in Quantification
  Applications (Preliminary Draft)}. In \bibinfo{booktitle}{\emph{Proceedings
  of the 8th International Workshop on Information Filtering and Retrieval
  (DART 2014)}}. \bibinfo{address}{Pisa, {IT}}, \bibinfo{pages}{1--11}.
\newblock


\bibitem[\protect\citeauthoryear{Esuli and Sebastiani}{Esuli and
  Sebastiani}{2015}]%
        {Esuli:2015gh}
\bibfield{author}{\bibinfo{person}{Andrea Esuli} {and}
  \bibinfo{person}{Fabrizio Sebastiani}.} \bibinfo{year}{2015}\natexlab{}.
\newblock \showarticletitle{Optimizing Text Quantifiers for Multivariate Loss
  Functions}.
\newblock \bibinfo{journal}{\emph{ACM Transactions on Knowledge Discovery and
  Data}} \bibinfo{volume}{9}, \bibinfo{number}{4} (\bibinfo{year}{2015}),
  \bibinfo{pages}{Article 27}.
\newblock
\urldef\tempurl%
\url{https://doi.org/10.1145/2700406}
\showDOI{\tempurl}


\bibitem[\protect\citeauthoryear{Forman}{Forman}{2008}]%
        {Forman:2008kx}
\bibfield{author}{\bibinfo{person}{George Forman}.}
  \bibinfo{year}{2008}\natexlab{}.
\newblock \showarticletitle{Quantifying counts and costs via classification}.
\newblock \bibinfo{journal}{\emph{Data Mining and Knowledge Discovery}}
  \bibinfo{volume}{17}, \bibinfo{number}{2} (\bibinfo{year}{2008}),
  \bibinfo{pages}{164--206}.
\newblock
\urldef\tempurl%
\url{https://doi.org/10.1007/s10618-008-0097-y}
\showDOI{\tempurl}


\bibitem[\protect\citeauthoryear{Gao and Sebastiani}{Gao and
  Sebastiani}{2016}]%
        {Gao:2016uq}
\bibfield{author}{\bibinfo{person}{Wei Gao} {and} \bibinfo{person}{Fabrizio
  Sebastiani}.} \bibinfo{year}{2016}\natexlab{}.
\newblock \showarticletitle{From Classification to Quantification in Tweet
  Sentiment Analysis}.
\newblock \bibinfo{journal}{\emph{Social Network Analysis and Mining}}
  \bibinfo{volume}{6}, \bibinfo{number}{19} (\bibinfo{year}{2016}),
  \bibinfo{pages}{1--22}.
\newblock
\urldef\tempurl%
\url{https://doi.org/10.1007/s13278-016-0327-z}
\showDOI{\tempurl}


\bibitem[\protect\citeauthoryear{Gonz{\'{a}}lez, Casta{\~{n}}o, Chawla, and del
  Coz}{Gonz{\'{a}}lez et~al\mbox{.}}{2017}]%
        {Gonzalez:2017it}
\bibfield{author}{\bibinfo{person}{Pablo Gonz{\'{a}}lez},
  \bibinfo{person}{Alberto Casta{\~{n}}o}, \bibinfo{person}{Nitesh~V. Chawla},
  {and} \bibinfo{person}{Juan~Jos{\'{e}} del Coz}.}
  \bibinfo{year}{2017}\natexlab{}.
\newblock \showarticletitle{A Review on Quantification Learning}.
\newblock \bibinfo{journal}{\emph{Comput. Surveys}} \bibinfo{volume}{50},
  \bibinfo{number}{5} (\bibinfo{year}{2017}), \bibinfo{pages}{74:1--74:40}.
\newblock
\urldef\tempurl%
\url{https://doi.org/10.1145/3117807}
\showDOI{\tempurl}


\bibitem[\protect\citeauthoryear{Joachims}{Joachims}{2005}]%
        {Joachims05}
\bibfield{author}{\bibinfo{person}{Thorsten Joachims}.}
  \bibinfo{year}{2005}\natexlab{}.
\newblock \showarticletitle{A support vector method for multivariate
  performance measures}. In \bibinfo{booktitle}{\emph{Proceedings of the 22nd
  International Conference on Machine Learning (ICML 2005)}}.
  \bibinfo{address}{Bonn, {DE}}, \bibinfo{pages}{377--384}.
\newblock


\bibitem[\protect\citeauthoryear{Maas, Daly, Pham, Huang, Ng, and Potts}{Maas
  et~al\mbox{.}}{2011}]%
        {Maas2011}
\bibfield{author}{\bibinfo{person}{Andrew~L. Maas}, \bibinfo{person}{Raymond~E.
  Daly}, \bibinfo{person}{Peter~T. Pham}, \bibinfo{person}{Dan Huang},
  \bibinfo{person}{Andrew~Y. Ng}, {and} \bibinfo{person}{Christopher Potts}.}
  \bibinfo{year}{2011}\natexlab{}.
\newblock \showarticletitle{Learning word vectors for sentiment analysis}. In
  \bibinfo{booktitle}{\emph{Proceedings of the 49th Annual Meeting of the
  Association for Computational Linguistics (ACL 2011)}}.
  \bibinfo{address}{Portland, {US}}, \bibinfo{pages}{142--150}.
\newblock


\bibitem[\protect\citeauthoryear{Platt}{Platt}{2000}]%
        {Platt:2000fk}
\bibfield{author}{\bibinfo{person}{John~C. Platt}.}
  \bibinfo{year}{2000}\natexlab{}.
\newblock \showarticletitle{Probabilistic outputs for support vector machines
  and comparison to regularized likelihood methods}.
\newblock In \bibinfo{booktitle}{\emph{Advances in Large Margin Classifiers}},
  \bibfield{editor}{\bibinfo{person}{Alexander Smola}, \bibinfo{person}{Peter
  Bartlett}, \bibinfo{person}{Bernard Sch{\"o}lkopf}, {and}
  \bibinfo{person}{Dale Schuurmans}} (Eds.). \bibinfo{publisher}{The MIT
  Press}, \bibinfo{address}{Cambridge, MA}, \bibinfo{pages}{61--74}.
\newblock


\bibitem[\protect\citeauthoryear{Saerens, Latinne, and Decaestecker}{Saerens
  et~al\mbox{.}}{2002}]%
        {Saerens:2002uq}
\bibfield{author}{\bibinfo{person}{Marco Saerens}, \bibinfo{person}{Patrice
  Latinne}, {and} \bibinfo{person}{Christine Decaestecker}.}
  \bibinfo{year}{2002}\natexlab{}.
\newblock \showarticletitle{Adjusting the Outputs of a Classifier to New a
  Priori Probabilities: A Simple Procedure}.
\newblock \bibinfo{journal}{\emph{Neural Computation}} \bibinfo{volume}{14},
  \bibinfo{number}{1} (\bibinfo{year}{2002}), \bibinfo{pages}{21--41}.
\newblock


\end{thebibliography}

\end{document}